\documentclass[wcp]{jmlr}

\usepackage{longtable}

\usepackage{booktabs}

\usepackage{lineno}

\makeatletter
\let\Ginclude@graphics\@org@Ginclude@graphics 
\makeatother

\jmlrvolume{222}
\jmlryear{2023}
\jmlrworkshop{ACML 2023}

\usepackage{caption}
\usepackage{subcaption}
\newtheorem{prop}{Proposition}
\usepackage{comment}
\newcommand{\pch}[1]{\textcolor{black}{#1}}
\newcommand{\ho}[1]{\textcolor{black}{#1}}
\newcommand{\icwu}[1]{\textcolor{black}{#1}}
\newcommand{\ds}[1]{\textcolor{black}{#1}}
\usepackage{algorithm}
\usepackage{algorithmicx}%
\usepackage{algpseudocode}%
\usepackage{multirow}

\title[Towards Human-Like RL]{Towards Human-Like RL: Taming Non-Naturalistic Behavior in Deep RL via Adaptive Behavioral Costs in 3D Games}

\author{\Name{Kuo-Hao Ho}$^*$ \Email{lukewayne123.cs05@nycu.edu.tw}\\
\Name{Ping-Chun Hsieh}$^*$ \Email{pinghsieh@nycu.edu.tw}\\
\Name{Chiu-Chou Lin} \Email{dsobscure@outlook.com}\\
\Name{You-Ren Luo} \Email{s9016010@gmail.com}\\
\Name{Feng-Jian Wang} \Email{fjwang@cs.nctu.edu.tw}\\
\Name{I-Chen Wu} \Email{icwu@cs.nctu.edu.tw}\\
\addr Department of Computer Science, National Yang Ming Chiao Tung University, No. 1001, Daxue Road, Taiwan
}

\editors{Berrin Yan{\i}ko\u{g}lu and Wray Buntine}

\begin{document}

\maketitle
\def\thefootnote{*}\footnotetext{These authors contributed equally to this work}

\abstract{In this paper, we propose a new approach called Adaptive Behavioral Costs in Reinforcement Learning (ABC-RL) for training a human-like agent with competitive strength. While deep reinforcement learning agents have recently achieved superhuman performance in various video games, some of these unconstrained agents may exhibit actions, such as shaking and spinning, that are not typically observed in human behavior, resulting in peculiar gameplay experiences. To behave like humans and retain similar performance, ABC-RL augments behavioral limitations as cost signals in reinforcement learning with dynamically adjusted weights. Unlike traditional constrained policy optimization, we propose a new formulation that minimizes the behavioral costs subject to a constraint of the value function. By leveraging the augmented Lagrangian, our approach is an approximation of the Lagrangian adjustment, which handles the trade-off between the performance and the human-like behavior. 
Through experiments conducted on 3D games in DMLab-30 and Unity ML-Agents Toolkit, we demonstrate that ABC-RL achieves the same performance level while significantly reducing instances of shaking and spinning. These findings underscore the effectiveness of our proposed approach in promoting more natural and human-like behavior during gameplay.
}

\keywords{Deep reinforcement learning, Human-like agent, Constrained policy optimization}




\section{Introduction}
\label{sec:introduction}

In recent years, many deep reinforcement learning (DRL) agents have achieved superhuman performance in many games, such as Agent57 \citep{Agent57} for Atari games, AlphaStar for StarCraftII \citep{Vinyals2019GrandmasterLI}, OpenAI Five for Dota2 \citep{OpenAIFive}, and For The Win (FTW) agent for Quake III Arena \citep{Jaderberg2019HumanlevelPI}. 
%
Although many agents can achieve good performance, they do not always behave like humans. 
One example is a high number of action per minute (APM). Without a limitation of APM, the peak APM in AlphaStar \citep{alphastarblog,Vinyals2019GrandmasterLI} sometimes reached 900 and even 1500, far above any human players, at most 600. For this problem, they limited the APM of AlphaStar
for a fair comparison to human players. 
Another example is frequent actions of shaking and spinning in 3D games, as observed in Banana Collector, a game in Unity Machine Learning Agents Toolkit, denoted as ML-Agents Toolkit \citep{Unity}. 
We noticed that the agent achieved a commendable score; however, its actions frequently exhibited shaking or spinning, which resulted in peculiar actions that were unsettling for human players. 
The phenomenon of shaking was also observed in the demo video\footnote{https://deepmind.com/blog/article/capture-the-flag-science} of FTW agent \citep{Jaderberg2019HumanlevelPI}.

\ho{
\ds{
As we reviewed the aforementioned studies with notable performance, some researchers are further exploring methods for demonstrating human-like behavior in games \citep{DiscussHuman-likeAgents_momennejad2023rubric,DiscussHuman-likeAgents_firai/NajarC21}. For example, \citet{Human-likeNavigationTuringTest_DevlinGMRZCLSH21} and \citet{Human-likeNavigationTuringTest_ZunigaMLRGMBSSC22} proposed different approaches to evaluate the human-likeness of navigation behavior in video games. In addition, \citet{HumanLikeNavigationAgent_MilaniJMGRSCFDH23} developed a human-like navigation agent using reward-shaping techniques \citep{reward-shaping_RosenfeldCTK18}. Also, in the work by \citet{HumanLikeDiplomacyGame_JacobWFLHBAB22}, a regret minimization algorithm was employed for search, demonstrating a policy that matches the human prediction and performs strongly in the no-press Diplomacy game. Other approaches incorporate human data to achieve human-like behaviors \citep{Human-likeCar-following, Human-LikeCollisionAvoidance, LearningHumanLikeStype_aiide/WoillemontLC22}.
These studies have comprehensively advanced the development of human-like agents in various games from different perspectives.
}}

\ho{
\ds{
To leverage human data, a straightforward approach is to employ imitation learning (IL) using human demonstrations. Techniques such as behavior cloning and generative adversarial imitation learning (GAIL) can be utilized for this purpose \citep{GAIL}.
In addition to directly learning from human demonstrations, it is possible to incorporate techniques from offline reinforcement learning to ensure agents do not deviate significantly from the provided demonstrations. Examples of such techniques include Conservative Q-Learning \citep{ConservativeQ-Learning} or Extreme Q-Learning \citep{ExtremeQ-Learning}.
While these IL approaches can capture human-like behavior, they usually heavily rely on demonstrations, thereby requiring sufficient samples for training. The performance of agents is bounded by these demonstrations. Thus, we explore another avenue besides imitation learning.
}
}



\ds{
In this paper, we employ constrained reinforcement learning and intuitive human-like metrics to achieve human-like behavior. These metrics are associated with biological constraints, such as behaviors like shaking or spinning in 3D games.
}
\ds{
The term \textit{Biological Constraints} was introduced by \citet{Biological_constraint} to define the physical limitations influencing human-like behavior in the context of games. They proposed methods to train an agent for the game Mario using these constraints. These constraints encompass sensory error, perceptual and motion delay, physical fatigue, and the balance between repetition and novelty.
}
\ds{
The limitation of APM described above can be viewed as one example of the biological constraints. \citet{Biological_constraint} also proposed some methods to address these constraints. For instance, they suggest adding noises to the game states to account for sensory error and delaying the input frame for perceptual and motion delay. However, some of these methods require internal information. Therefore, achieving human-like behavior through end-to-end training remains challenging. Thus, for simplicity, we focus on the issue of frequent actions of shaking and spinning as previously described.
}

To solve the issue of frequent actions of shaking and spinning, we first propose a metric to indicate the cost of such behaviors due to biological constraints, called {\em behavioral costs} in this paper. The metric of behavioral costs can be defined by game designers or biologists, e.g., the costs go high for frequent shaking and spinning.

Second, letting the behavioral costs as a negative reward to discourage non-human behaviors, our goal is to find a reinforcement learning (RL) policy that minimizes behavioral costs while achieving sufficiently high total rewards. Thus, the problem can be viewed as a kind of constrained policy optimization (CPO) problem. Notably, in most traditional CPOs, the target is to find a policy that maximizes the total rewards subject to the given constraints, such as the limitation of behavioral costs. Unlike traditional constrained policy optimization, we propose a new formulation that minimizes the behavioral costs subject to a constraint of the value function. 

Third, leveraging augmented Lagrangian, our approach is to transfer the problem into unconstrained policy optimization. Primal–dual optimization (PDO) is a widely used approach for Lagrangian, but its solution for the primal problem must satisfy its dual constraint. In this paper, the agent is allowed with a few violations as the fact that a human player sometimes shakes or spins. Thus, we propose a new approximation approach, called Adaptive Behavioral Costs in Reinforcement Learning (ABC-RL), to adjust rewards dynamically, so that we can obtain high performance while making the action of shaking and spinning less frequent.

Finally, we justify the ABC-RL approach by experimenting with training agents in Banana Collector and DMLab-30~\citep{DMLab}. The experiments show that the agents trained by our approach greatly reduce the numbers of shaking and spinning while preserving the same performance level. In addition, we also compare our agents to human players. In this experiment, while outperforming human players, our agents take slightly more spinning actions than human players do, but less shaking actions. Our conjecture about this is that human players tend not to spin than to shake.

\section{Preliminaries} 
\label{sec:preliminaries}

In this section, we introduce two preliminaries and its related works, including reinforcement learning (RL) and constrained policy optimization(CPO).

\subsection{Reinforcement Learning}
\label{sec:MDP}
A RL problem is usually formulated as a Markov Decision Process (MDP), defined by a tuple (\textit{S}, \textit{A}, \textit{R}, \textit{P}, \textit{$\gamma$}), where
\textit{S} is the finite set of the states, 
\textit{A} is the set of the available actions, 
\textit{R}: $\textit{S} \times \textit{A} \rightarrow \mathbb{R}$ is the reward function, 
\textit{P}: $\textit{S} \times \textit{A} \rightarrow \textit{S} \mapsto [0,1]$ is the transition probability function and 
\textit{$\gamma$} is the discounted factor. At each time step \textit{t}, the agent decides an action $a_t \in A$ by its policy $\pi(a_t|s_t)$ at state $s_t \in S$; then, by applying the action $a_t$, the state of the environment is changed from $s_t$ to $s_{t+1}$ according to the environment transition probability function $P$ and the agent receives a feedback reward $r_t \in R$. The above process continues until the agent reaches a terminal state. 

The goal for a RL agent is to learn a policy \textit{$\pi_\theta$} parametrized by $\theta$ from some parameter set $\Theta$ for maximizing its performance measure, called objective function $J(\pi_\theta)$. 
For a fixed policy, the objective function can be represented as the expected total infinite horizon discounted rewards, $J_v(\pi_\theta)= \mathbb{E}_{\tau\sim\pi_\theta}[\sum_{t=0}^{\infty}\gamma^tr_t]$, where $\tau=(s_0,a_0,s_1,a_1,\dots)$ is a trajectory played by this fixed policy $\pi_\theta$ and $\tau\sim\pi_\theta$ presents the distributions of the states and actions over the trajectories depending on $\pi_\theta$. 
We denote a return $G(\tau)$ as the total discounted rewards of a trajectory; thus, the objective function can be rewritten as  $J_v(\pi_\theta)=\mathbb{E}_{\tau\sim\pi_\theta}[G(\tau)]$. 
It is considered as an optimization problem that finding the parameters $\theta$ for a policy to maximize the objective function $J_v(\pi_\theta)$, as Eq.~\ref{eq:optimization_problem}. This optimization problem is a policy optimization problem, which is usually solved by policy gradient approaches
~\citep{Sutton1988ReinforcementLA}.
\begin{equation} 
    \label{eq:optimization_problem}
    \max_{\theta}J_v(\pi_\theta)=\max_{\theta}\mathbb{E}_{\tau\sim\pi_\theta}[\sum_{t=0}^{\infty}\gamma^tr_t]
\end{equation} 

Since deep Q-learning network (DQN) \citep{Mnih2015HumanlevelCT} had great success in Atari games, it has been more popular to use neural networks to parameterize the agent's policy. Researchers have 
\ds{improved}
the 
\ds{performance}
with many techniques, such as 
\ds{refining the}
value function \citep{DDQN,DuelingDQN} and 
\ds{accelerating}
the training process with a distributed system \citep{A3C,APEX,R2D2}. However, the target of those approaches is for high performance, such as getting more game scores. Without extra limitations, the behaviors of the agent are not guaranteed to be human-like. Inspired by \ds{the work of} \citet{Biological_constraint} 
, which introduced "biological constraints" to make RL agent more human-like, the behavioral limitations are introduced as constraints to training a human-like agent. 

\subsection{Constrained Policy Optimization}
Constrained Markov Decision Processes (CMDP) \citep{CMDP} is an extension of MDP with introducing constraints that the policy must fulfill. The general form is 
\begin{equation}
    \label{eq:org CPO}
    \max_{\theta}J_v(\pi_\theta)\mbox{, subject to }J_c(\pi_\theta) \leq \tau,
\end{equation}
where $\tau$ is the threshold for the constraints and $J_c(\pi_\theta)$ is the constrained costs, such as discounted total costs $J_c(\pi_\theta) = \mathbb{E}_{\tau\sim\pi_\theta}[\sum_{t=0}^{\infty}\gamma^t C(t)]$, where $C(t)$ is the behavioral costs signal at time $t$. 

CMDP is one of the approaches for Safe RL, \ds{which is} RL with safety constraints 
\ds{to avoid} dangerous actions
\ds{, and it is also considered as constrained RL \citep{SafeRLsurvey}.}
The Lagrangian approach is a widely used technique for solving CMDP problems, such as primal–dual optimization (PDO) \citep{PDO-CMDP2003,PDO-CMDP2017} and multi-timescale actor-critic approach \citep{AC-CMDP2005,RCPO-CPO2019}. Recently, \citet{Achiam2017ConstrainedPO} introduced the Constrained Policy Optimization (CPO), a policy search algorithm for CMDPs. Their experiments in control tasks showed CPO was able to maximize the return while approximately satisfying constraints. But, CPO only handles single constraint, which is not suitable for behavioral costs. There are many approaches for constrained RL, such as Interior-point Policy Optimization (IPO) \citep{IPO-CPO2020}, which uses first-order policy optimization method, and Projection-Based Constrained Policy Optimization (PCPO) \citep{Projection-CPO2020}, which finds a policy that satisfies the constraints by projections. However, we consider both approaches are not suitable for our problem. IPO is not able to violate the constraints during training while our approach gives some freedom in taking non-human-like behavior. 
\ds{Also,} we think computing for projections is too complex in PCPO. Thus, we provide a novel view to solve CPO. 

\section{Methods}
\label{sec:methods}
\subsection{\pch{Human-Like Reinforcement Learning via Adaptive Behavioral Costs}}
\pch{In this section, we present the general ABC-RL framework to attain competitive performance as well as human-like behavior. 
To achieve both targets simultaneously, we resort to the paradigm of constrained policy search, i.e.\ds{,} formulate one of the targets as the objective function while taking the other into account through constraints.
In the typical constrained policy optimization problem \citep{Achiam2017ConstrainedPO}, 
\ds{a}
RL agent is trained by maximizing the expected total return subject to the constraints of the total expected costs. 
While being an attractive solution, this formulation is not directly applicable to human-like reinforcement learning for the following reason: To learn a competitive policy, the agent may still require some freedom in taking non-human-like behavior. For example, in Banana Collector, the agent needs to make a minimum level of shaking to effectively collect bananas.
However, without experiments, it is usually impractical for the researchers to configure the cost constraints at an appropriate level.
Without the well-configured constraints, the training progress may stall, or the performance of the learned policy can be rather poor.}

\pch{To address this issue, we propose the following formulation for policy optimization:}
\begin{equation}
    \label{eq:BCPO-opt-1}
    \min_{\theta}J_c(\pi_\theta)\mbox{, subject to }J_v(\pi_\theta) \geq V_{\text{th}},
\end{equation} 
where $J_c(\pi_\theta) \geq 0$ presents the total discounted costs, $J_v(\pi_\theta)$ is the expected value of the policy $\pi_\theta$ and $V_{\text{th}}$ is a threshold of the value.
\pch{The above formulation is more natural to human-like reinforcement learning since it aims to minimize unnecessary non-human-like behavior while guaranteeing sufficiently high performance.}
\pch{Note that in the above formulation\ds{,} we do not assume any specific forms of the behavioral costs, which are to be designed by the researchers for their own purposes. As will be seen in 
\ds{Section~}\ref{sec: shaking and spinning cost} and \ref{sec:experiments}, we implement two types of behavioral costs to evaluate the proposed framework.}
One way to define the behavioral costs signal is given in 
\ds{Section~}\ref{sec: shaking and spinning cost}, and also followed in our experiments in Section \ref{sec:experiments}.
\ho{
$V_{\text{th}}$ could be a hyperparameter or be assigned as a fraction of the maximum historical value ($V_{\max}$) in the unconstrained case.
\ds{Equation~}\ref{eq:BCPO-opt-1}
indicates the behavioral costs are going to decrease after the performance reaches a certain threshold.
In practice, we propose to select $V_{\text{th}}$ based on the achievable performance of the unconstrained problem. 
In our experiments, it is suggested that $V_{\text{th}}=0.8\cdot V_{\max}$.
}

To solve the optimization problem of the ABC-RL, one typical approach is to approximately solve Equation~\ref{eq:BCPO-opt-1} by converting the constrained problem into a single unconstrained problem via the \textit{quadratic penalty method}.
Specifically, the quadratic penalty method relaxes the constraint by adding to the original objective a high penalty that reflects the constraint violation.
Despite its simplicity, the penalty method is known to suffer from the ill-conditioning issue \citep{bertsekas1997nonlinear}.
To address this issue, we adopt the augmented Lagrangian approach \citep{bertsekas1997nonlinear}, instead of the quadratic penalty method. 
To construct the augmented Lagrangian of Equation~\ref{eq:BCPO-opt-1}, we first introduce a dummy scalar variable $z\in\mathbb{R}$ such that the constraint can be rewritten as $V_{\text{th}}-J_v(\pi_{\theta})+z^2 =0$.
The augmented Lagrangian associated with Equation~\ref{eq:BCPO-opt-1} is defined as
\begin{equation}
    L(\theta,z;\lambda,\mu):=J_c(\pi_\theta) + \lambda(V_{\text{th}} - J_v(\pi_\theta) + z^2) + \frac{\mu}{2}(V_{\text{th}} - J_v(\pi_\theta) + z^2)^2,
    \label{eq:augemented Lagrangian}
\end{equation}
where $\lambda$ is the Lagrange multiplier, and $\mu$ is the penalty parameter.
Subsequently, Equation~\ref{eq:BCPO-opt-1} can be approximately solved as follows:
\begin{equation}
    \label{eq:BCPO-opt-2}
    \min_{\theta\in \Theta,z\in\mathbb{R}}L(\theta,z;\lambda,\mu).
\end{equation}
Since Equation~\ref{eq:BCPO-opt-2} is equivalent to the double minimization problem $\min_{\theta\in \Theta}\min_{z\in\mathbb{R}} L(\theta,z;\lambda,\mu)$, we can simplify the procedure by first handling the inner minimization over $z$.

\begin{prop}
\label{prop:equivalent problem}
\normalfont The solution of $\theta$ to Equation~\ref{eq:BCPO-opt-2} is the same as that to the following problem:
\begin{equation}
    \label{eq:BCPO-opt-2B}
    \min_{\theta\in\Theta}\hspace{3pt}J_c(\pi_\theta) + \frac{1}{2\mu}\big((\max\{0, \lambda + \mu(V_{\text{th}} - J_v(\pi_\theta))\})^2-\lambda^2\big). 
\end{equation}
\end{prop}
The proof of Proposition \ref{prop:equivalent problem} is as follows:
\begin{align}
    \label{eq:prop1}
    &\hspace{6pt}\min_{\theta\in \Theta,z\in\mathbb{R}} L(\theta,z;\lambda,\mu)\\
    =&\hspace{6pt}\min_{\theta\in \Theta}\min_{z\in\mathbb{R}}J_c(\pi_\theta) + \lambda(V_{\text{th}} - J_v(\pi_\theta) + z^2) + \frac{\mu}{2}(V_{\text{th}} - J_v(\pi_\theta) + z^2)^2\label{eq:prop1-z}\\
    =&\hspace{6pt}\min_{\theta\in \Theta}\min_{y\geq0}J_c(\pi_\theta) + \lambda(V_{\text{th}} - J_v(\pi_\theta) + y) + \frac{\mu}{2}(V_{\text{th}} - J_v(\pi_\theta) + y)^2\label{eq:prop1-y}\\
    =&\hspace{6pt}\min_{\theta\in \Theta}J_c(\pi_\theta) + \lambda(\max\{V_{\text{th}} - J_v(\pi_\theta), -\frac{\lambda}{\mu}\}) + \frac{\mu}{2}(\max\{V_{\text{th}} - J_v(\pi_\theta), -\frac{\lambda}{\mu}\})^2\label{eq:prop1-y-analysis}\\
    =&\hspace{6pt}\min_{\theta}J_c(\pi_\theta) + \frac{1}{2\mu}(\max\{0, \lambda + \mu(V_{\text{th}} - J_v(\pi_\theta))\})^2-\lambda^2)\label{eq:prop1-y-solution}.
\end{align}
Equation~\ref{eq:prop1-y} is replaced the variable $z$ in Equation~\ref{eq:prop1-z} with a non-negative variable $y$. As Equation~\ref{eq:prop1-y} is a quadratic function of $y$, we can simplify Equation~\ref{eq:prop1-y} to Equation~\ref{eq:prop1-y-analysis} by $y=0$ or $y=-\lambda/\mu-(V_{\text{th}} - J_v(\pi_\theta))$. By analyzing the value of the maximum term ($\max\{V_{\text{th}} - J_v(\pi_\theta), -\lambda/\mu\}=V_{\text{th}} - J_v(\pi_\theta)$ or $-\lambda/\mu$), the solution of $\theta$ to Equation~\ref{eq:prop1-y-analysis} is equal to that to Equation~\ref{eq:prop1-y-solution}. 

Given that the parameter $\theta$ is updated iteratively during training, we consider an iterative procedure to implement Equation~\ref{eq:BCPO-opt-2B} as follows:
\ho{
If the old policy and the new policy are close enough, we can apply the linear approximation to represent the first-order differential for the second term in Equation~\ref{eq:BCPO-opt-2B}
}.
Specifically, we consider
\pch{
\begin{align}
    \min_{\theta\in\Theta}&\hspace{6pt}J_c(\pi_\theta) - (\max\{0,\lambda+\mu(V_{\text{th}} - J_v({\pi_{\theta_{\text{old}}}}))\})J_v(\pi_\theta)\label{eq:BCPO-opt-3}\\
    \mbox{subject to }&
    \hspace{6pt}\lvert{J_v(\pi_\theta)}-{J_v(\pi_{\theta_{\text{old}}})} \rvert\leq \varepsilon,\label{eq:BCPO-opt-3 constraint}
\end{align}
where Equation~\ref{eq:BCPO-opt-3 constraint} is the proximity constraint.
In practice, there are various ways to ensure that Equation~\ref{eq:BCPO-opt-3 constraint} is satisfied, such as the inclusion of a KL divergence constraint \citep{Achiam2017ConstrainedPO} or using the clipped objective function \citep{PPO}.
}
Let $\lambda_t$ denote the Lagrange multiplier for the $t$-th episode.
Following the augmented Lagrangian  \citep{bertsekas1997nonlinear}, the Lagrange multiplier $\lambda$ will be updated at the end of each episode as $\lambda_{t+1}=\max\{0,\lambda_t+\mu\big(V_{\text{th}}-J_v({\pi_\theta})\big)\}$.
To make Equation~\ref{eq:BCPO-opt-3}-\ref{eq:BCPO-opt-3 constraint} more compatible with the convention of maximizing total return in RL, we consider the following equivalent maximization problem as
\begin{align}
    \max_{\theta\in\Theta}&\hspace{6pt}J_v({\pi_\theta}) - \frac{1}{\max\{0,\lambda+\mu\big(V_{\text{th}} - J_v({\pi_{\theta_{\text{old}}}})\big)\}}J_c(\pi_\theta)\label{eq:BCPO-opt-3B} 
    \\
    \mbox{subject to }& \hspace{6pt}\lvert{J_v(\pi_\theta)}-{J_v(\pi_{\theta_{\text{old}}})} \rvert\leq \varepsilon.
   \label{eq:BCPO-opt-3B constraint} 
\end{align}
\ho{
Equation~\ref{eq:BCPO-opt-3}-\ref{eq:BCPO-opt-3 constraint} and Equation~\ref{eq:BCPO-opt-3B}-\ref{eq:BCPO-opt-3B constraint} are equivalent if $\lambda +\mu(V_{\text{th}}-J_v({\pi_{\theta_{\text{old}}}})) > 0$. If $\lambda +\mu(V_{\text{th}}-J_v({\pi_{\theta_{\text{old}}}})) \leq 0$, this implies that $J_v({\pi_{\theta_{\text{old}}}})$ is sufficiently large (i.e., $J_v({\pi_{\theta_{\text{old}}}})\geq V_{\text{th}}+\lambda/\mu$) such that (i) the objective in Equation~\ref{eq:BCPO-opt-3} reduces to the total expected behavioral cost $J_c({\pi_\theta})$ and (ii) the weight of $J_c({\pi_\theta})$ in Equation~\ref{eq:BCPO-opt-3B} shall be negative infinity. In practice, we can avoid zero denominator in Equation~\ref{eq:BCPO-opt-3B} by either clipping the value of $\frac{1}{(\max\{0, \lambda+\mu(V_{\text{th}}-J_v({\pi_{\theta_{\text{old}}}}))\})}$ or selecting a small $\delta>0$ and use $\frac{1}{(\max\{\delta, \lambda+\mu(V_{\text{th}}-J_v({\pi_{\theta_{\text{old}}}}))\})}$.
}

Note that Equation~\ref{eq:BCPO-opt-3B} suggests a simple and intuitively appealing way to implement the ABC-RL framework.
That is, the term $({\lambda+\mu\big(V_{\text{th}} - J_v({\pi_{\theta_{\text{old}}}})\big)})^{-1}$ can be viewed as a weight for tuning the penalty induced by the cost signals, and this weight is determined automatically based on the current learning progress reflected by $V_{\text{th}} - J_v({\pi_{\theta_{\text{old}}}})$. 
Moreover, we discuss the following two regimes: 
\begin{itemize}
    \item Low-penalty regime: If $V_{\text{th}}$ is much larger than $J_v({\pi_{\theta_{\text{old}}}})$, the penalty weight is rather small, and the agent shall behave as if there is no behavioral cost. This regime guarantees effective learning during the initial training phase.
    \item High-penalty regime: If $V_{\text{th}} - J_v({\pi_{\theta_{\text{old}}}})$ is close to 0, then the penalty weight is roughly equal to $1/\lambda$. Hence, Equation~\ref{eq:BCPO-opt-3B} is reduced to the form of the standard Lagrangian.
\end{itemize}


Built on the above derivation of Equation~\ref{eq:BCPO-opt-3B}, we further discuss the practical consideration in the algorithm design:
First, as the true value of $J_v(\pi_{\text{old}})$ is not directly accessible and requires estimation, we use the average of total return over the previous episodes (denoted by $V_{\text{avg}}$) as an estimate of $J_v(\pi_{\theta_{\text{old}}})$, which makes the training progress more stable.
Second, in the low-penalty regime, the penalty weight is determined by multiple parameters, including $\lambda,\mu$, and $V_{\text{th}}$.
To guarantee a sufficiently small penalty weight for various environments in a more holistic manner, we propose to use the sigmoid function as the surrogate for the derived penalty weight in Equation~\ref{eq:BCPO-opt-3B}.
Specifically, the variant of Equation~\ref{eq:BCPO-opt-3B} with the sigmoid surrogate function is
\begin{equation}
    \label{eq:ABC-RL-opt}
    \max_{\theta\in\Theta}\hspace{6pt}J_v(\pi_\theta) - W\cdot \text{Sigmoid}\Big(\frac{V_{\text{avg}}-V_{\text{th}}}{h}\Big)\cdot J_c(\pi_\theta),
\end{equation}
\ho{
where $W$ denotes the maximum weight for the penalty, $h$ is the parameter that determines the slope of the sigmoid surrogate function and $V_{\text{avg}}$ is the average over the most recent $k$ episodes, where $k$ is 10 in our setting.
}
For ease of notation, we define $\Lambda:=W\cdot\text{Sigmoid}((V_{\text{avg}}-V_{\text{th}})/h)$ and call $\Lambda$ the \textit{adaptive behavioral weight}.
There are three salient features of the proposed sigmoid surrogate function:
(i) In the low-penalty regime, the value of the sigmoid surrogate is expected to be close to 0 since $V_{\text{avg}}$ is much smaller than the threshold $V_{\text{th}}$;
(ii) In the high-penalty regime, the surrogate function well mimics the behavior of the original penalty weight.
That is, given the property that $\frac{1}{{1+e^{-x}}}\approx\frac{1}{2-x}$, for all $x\ll 1$, 
we know that if $J_v(\pi_{\theta_{\text{old}}})$ is close to $V_{\text{th}}$, 
\begin{align}
    \label{eq:sigmoid approximation}
    \frac{1}{\lambda+\mu(V_{\text{th}} - J_v(\pi_{\theta_{\text{old}}}))}&=\frac{1}{\lambda}\frac{1}{1+\frac{\mu}{\lambda}(V_{\text{th}} - J_v(\pi_{\theta_{\text{old}}}))}\\
    & \approx\frac{2}{\lambda}\frac{1}{1+\exp\big({-\frac{2\mu}{\lambda}( J_v(\pi_{\theta_{\text{old}}})-V_{\text{th}} )}\big)}\\
    &= \frac{2}{\lambda}\text{Sigmoid}\Big(\frac{J_v(\pi_{\theta_{\text{old}}})-V_{\text{th}}}{\frac{\lambda}{2\mu}}\Big).
\end{align}
(iii) The sigmoid surrogate provides a smooth transition between the low-penalty and the high-penalty regimes.
Algorithm \ref{algo: ABC-RL} shows the pseudocode of the ABC-RL approach based on Equation~\ref{eq:ABC-RL-opt}.


\begin{algorithm}[h]
 \caption{Reinforcement Learning via Adaptive Behavioral Costs (ABC-RL)}
 \label{algo: ABC-RL}
\SetAlgoLined
 Initialize policy weigh $\theta$, replay buffer $B$\;\\
 Initialize constant $W$, $h$ and $V_{\text{th}}$\;\\
 \For{episode=1 to M} {
    Initialize state $s_0$\; \\
    \For{t=1 to T} {
        Apply the action $a_t$ from policy $\pi_\theta(s_t,a_t)$ and observe the reward $r_t$ and new state $s_{t+1}$\;\\
        Calculate the costs $C(t)$ \;\\
        Adjust reward $r_t'=r_t-\Lambda\cdot C(t)$, where $\Lambda=W\cdot \text{Sigmoid}(\frac{V_{\text{avg}}-V_{\text{th}}}{h})$\;\\
        Store transition $(s_t, a_t, r_t', s_{t+1})$\;\\
        Update $\theta$ by given RL method\;
    }
}
\end{algorithm}

On the other hand, the learning algorithm suggested by Equation~\ref{eq:BCPO-opt-3B} along with the corresponding Lagrange multiplier update is called AB-CPO in the rest of the paper.
Algorithm \ref{algo: AB-CPO} describes the pseudo code for Equation~\ref{eq:BCPO-opt-3B}.
In Algorithm \ref{algo: AB-CPO}, it needs to calculate $\lambda$ when the policy is stable. However, it is non-trivial to judge whether the policy is stable and what is the value of the penalty parameter $\mu$, which affects the Lagrange multiplier $\lambda$. 
\ho{
In practice, the policy is regarded stable if the average policy loss under the current training batch is below some threshold. 
Notably, as an alternative, using $V_{\text{avg}}$ as a surrogate for $J_v(\pi_\theta)$ could also mitigate the possible uncertainty and stochasticity in the observed total return.
}

\begin{algorithm}[h]
\caption{Adaptive Behavioral Constrained Policy Optimization (AB-CPO)}
\label{algo: AB-CPO}
\SetAlgoLined
 Initialize policy weigh $\theta$, replay buffer $B$, variables for each costs $\lambda_0$\;\\
 Initialize constant $\mu$ and $V_{\text{th}}$\;\\
 \For{episode=1 to M} {
    Initialize state $s_1$ \;\\
    \For{t=1 to T} {
        Apply action $a_t$ from policy $\pi_\theta(s_t,a_t)$ and observe reward $r_t$ and new state $s_{t+1}$\;\\
        Calculate costs $C(t)$ and adjust reward $r_t'=r_t-\Lambda\cdot C(t)$, where $\Lambda = \frac{1}{\lambda_t+\mu(V_{\text{th}}-J_v(\pi_\theta))}$\;\\
        Store transition $(s_t, a_t, r_t', s_{t+1})$\;\\
        Update $\theta$ by given policy gradient method\;
    }
    \If{the policy is stable} {
        Update the variables: $\lambda_{t+1}=\max\{0, \lambda_t+\mu(V_{\text{th}}-J_v(\pi_\theta))\}$\;
    }
}
\end{algorithm}

\subsection{Shaking and Spinning Cost} 
\label{sec: shaking and spinning cost}
\icwu{In the previous subsection, the behavioral costs signal is defined in a general aspect. For simplicity of analysis, we substantiate the proposed ABC-RL framework with two common types of behavioral costs signals in this subsection:
\begin{equation}
C(t) = C_{sh}(t)+\alpha C_{sp}(t),
\label{eq: cost signals}
\end{equation}
where $C_{sh}(t)$ denotes the shaking cost at time $t$, $C_{sp}(t)$ the spinning cost, and $\alpha$ is a hyperparameter for the importance between the penalties of the two costs.}  
Despite that the measures of human-like behavior can be rather subjective, we observe that excessive shaking and spinning are two major factors that make a well-trained RL agent appear non-human-like in 3D games. \icwu{This subsection presents one way to define the two costs.} 


\begin{figure}[htb!]
\centering
\begin{minipage}[b]{0.45\textwidth}
    \includegraphics[width=1\linewidth]{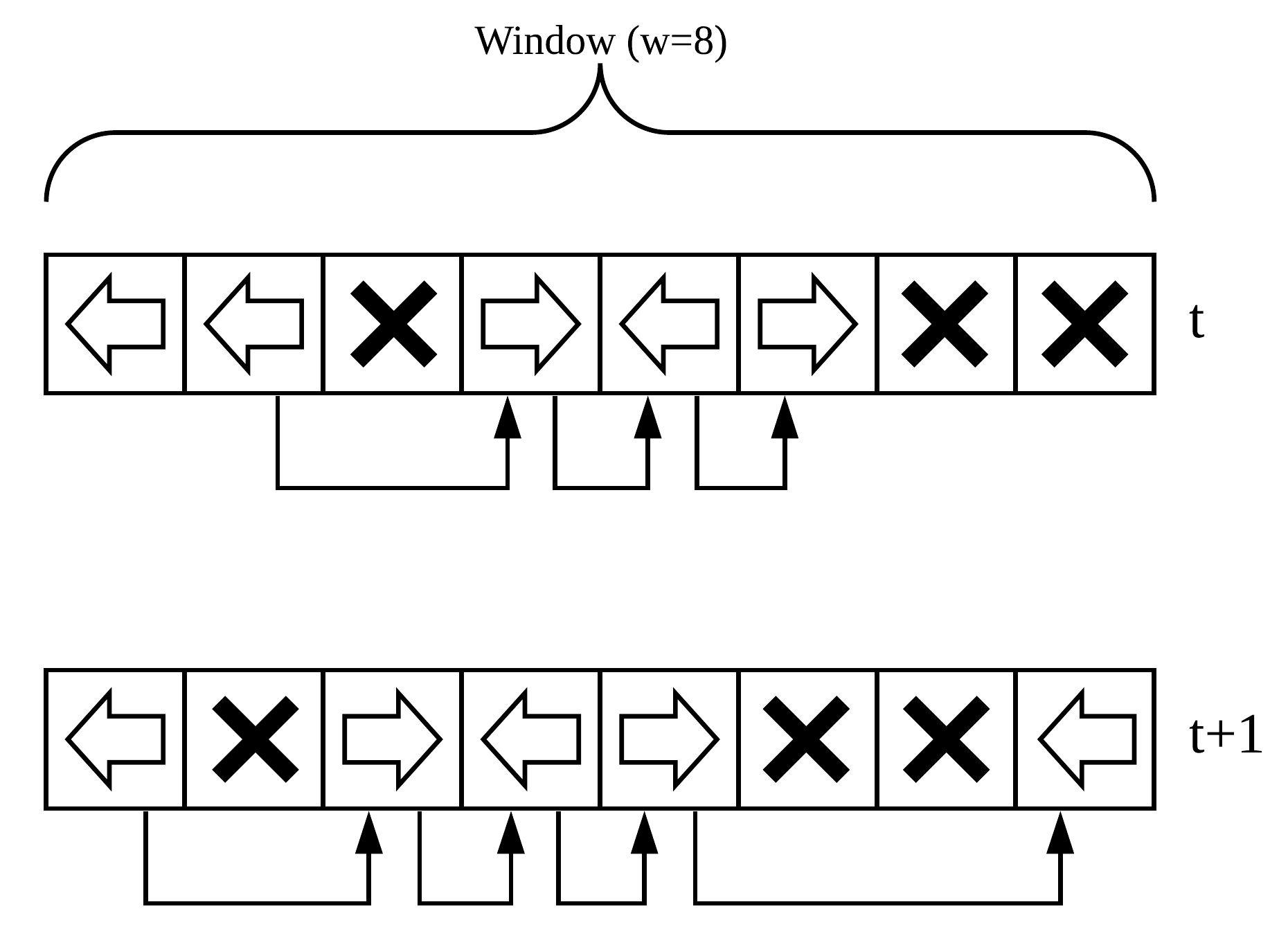}
    \centering
    (a) shaking cost
    \label{fig:shaking cost}
\end{minipage}
\begin{minipage}[b]{0.5\textwidth}
    \includegraphics[width=1\linewidth]{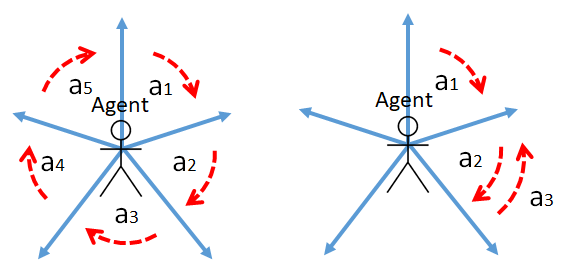}
    \centering
    (b) spinning cost
    \label{fig:spinning cost}
\end{minipage}
\caption{Shaking and spinning costs}
\label{fig: shake and spin }
\end{figure}

\pch{First, consider the shaking cost. In 3D games, the shaking behavior of an agent is usually recognized from the quick vibratory movements along a horizontal axis within a short period of time. In many games, shaking is an effective way for computer agents to search target, however, human players tend not to shake too often due to physical fatigue. In this paper, we quantify the amount of shaking as the number of changes in direction within a sliding time window of a fixed size $w\in\mathbb{N}$. Specifically, the shaking cost $C_{sh}(t)$ at time $t$ is defined as follows.}
\icwu{For simplicity, let horizontal actions include the three kinds of actions, move-left, move-right, and no operation (no-op). For a sequence of consecutive actions $<a_i, ..., a_j>$, if both $a_i$ and $a_j$ are opposite actions, namely move-left versus move-right, and all actions between the two are no-op, then count one for shaking. 
In a sequence of $w$ horizontal actions $<a_{t-w+1}, ..., a_t>$, the shaking cost $C_{sh}(t)$ is the total count divided by $w-1$ for normalization. Note that $w-1$ is the maximum count for shaking in a window. As illustrated by an example in Figure \ref{fig: shake and spin } (a), the shaking cost for the upper window of size $8$ is $3/7$, and that for the lower window is $4/7$.}

\icwu{Second, consider the spinning cost. In 3D games, shaking is also an effective way for computer agents to search target, and human players tend not to spin too often also due to physical fatigue. In this paper, we quantify the amount of spinning as the number of turning around and back to the original orientation. Specifically, the spinning cost $C_{sh}(t)$ at time $t$ is defined as follows.
Count one for spinning 
\ds{whenever} the agent turns one whole around, either left or right, and then face to the same orientation. As illustrated by an example in Figure \ref{fig: shake and spin } (b), the spinning cost is one for the left with the five actions $<a_1, ..., a_5>$, and nothing for the right. Note that for the left we count one more if the next five actions are the same as $<a_1, ..., a_5>$.}

\section{Experiments}
\label{sec:experiments}

Our experiments are run for a game on the ML-Agents Toolkit, called Banana Collector and some of the games on DMLab-30, described in Subsections  \ref{subsec:ml-agents experiment} and \ref{subsec:dmlab experiments} respectively. 

\subsection{ML-Agents Toolkit Experiment}
\label{subsec:ml-agents experiment}
Our version of ML-Agents Toolkit is 0.8.1. In the game of Banana Collector, the goal of the agent is to get as many yellow bananas (each with reward +1) as possible, while avoiding touching blue bananas (each with reward -1). The game is in a square area, and is ended after the agent takes 2000 steps. To make the area always exist some bananas, bananas will be refilled to the game periodically. For simplicity, the action space only includes the actions of moving and rotating. 
Other details are listed in Table \ref{tab:1}.

\begin{table}[h]
\caption{Environment settings}
\centering
\label{tab:1} 
\begin{tabular}{ |l|l| } 
\hline
Statement & Description \\
\hline
Input & 160x90 RGB image\\
\hline
\multirow{2}{5em}{Reward} & yellow banana: +1 \\ 
& blue banana: -1 \\ 
\hline
\multirow{2}{5em}{Action} & moving: forward, backward, none \\ 
& rotating: left, right, none \\ 
\hline
\multirow{4}{5em}{Game setting} & Game steps: 2000  \\ 
& Total agents: 4  \\ 
& FPS: 50  \\ 
& Frame skip: 4  \\ 
\hline
\end{tabular}
\end{table}

The ML-Agents Toolkit supports Proximal Policy Optimization (PPO) \citep{PPO} as a default implementation of the learning algorithm, which serves as the baseline agent (without behavioral costs) in this experiment. As this PPO version is implemented with a clipped surrogate objective version, it also enforces the proximity constraint Equation~\ref{eq:BCPO-opt-3 constraint}.
With behavioral costs, we train three agents, named ABC-RL, AB-CPO, and Const, based on Algorithm \ref{algo: ABC-RL} with the PPO implementation for update and with three different settings for the behavioral weight $\Lambda$ respectively. 
The agent named ABC-RL uses the same behavioral weight as in Algorithm \ref{algo: ABC-RL}, AB-CPO lets the weight $\Lambda$ follow the formula in Equation~\ref{eq:BCPO-opt-3B}, and Const fixes $\Lambda$ to constant 1. 
The behavioral costs follow the formula in Equation~\ref{eq: cost signals}, but let $\alpha=1$ for simplicity. 
Namely, $C(t)=C_{sh}(t)+C_{sp}(t)$. 
The window size for shaking costs is all set to be 8, i.e.\ds{,} $w=8$. 
\ho{Each experiment is run three times with one million steps each.}

\begin{figure}[htb!]
\begin{minipage}{0.55\linewidth}
  \includegraphics[width=1\linewidth]{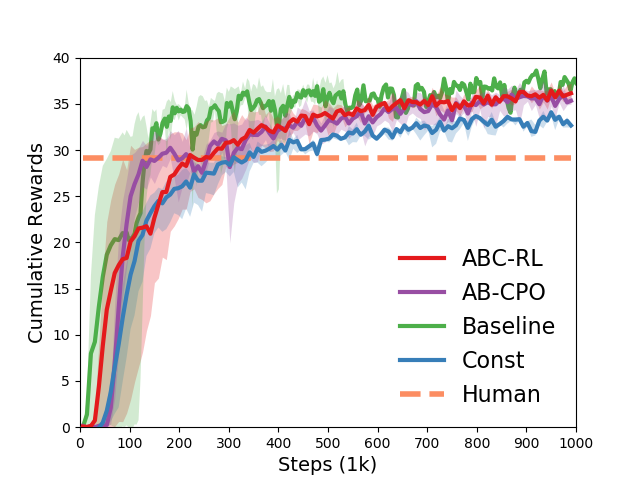}
  \centering
  (a) Cumulative rewards\label{fig:Unity/DataReward_steps-cpp}
\end{minipage}
\begin{minipage}{0.34\linewidth}
\includegraphics[width=\linewidth]{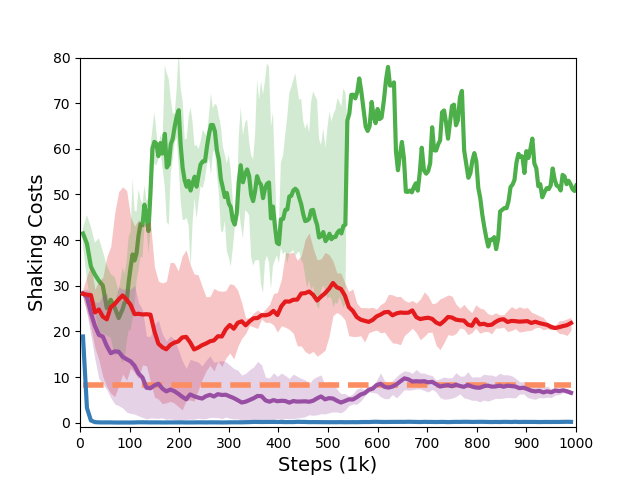}
\centering
(b) Shaking costs\label{fig:Unity/DataShakingCost_steps-cpp}
\includegraphics[width=\linewidth]{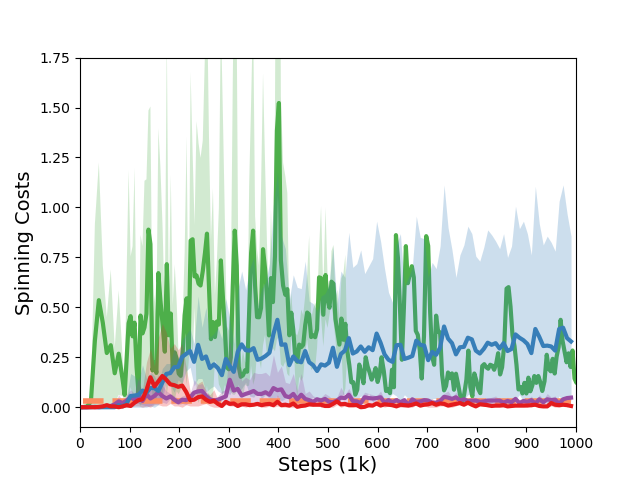}
\centering
(c) Spinning costs\label{fig:Unity/DataSpinningCost_steps-cpp}
\end{minipage}

\caption{Results in ML-Agents Toolkit with training steps}
\label{fig: ml-agnet result-cpo}
\end{figure}

The performances of the above four agents are shown in Figure \ref{fig: ml-agnet result-cpo}, including cumulative rewards, shaking costs, and spinning costs, where each curve presents the average value of three runs.
Without considering behavioral costs, the baseline performs the best for the cumulative rewards, while both shaking and spinning costs are apparently higher than others in most cases. 
With constant behavioral weight, the agent Const performs the worst clearly, while its shaking costs are the lowest among all agents and its spinning costs are apparently higher than ABC-RL and AB-CPO but comparable to the baseline near the end of the training. 
\ho{
The reason for the nearly zero shaking costs under Const is that the value of the shaking cost dominates the objective.
Thus, the agent Const learns to prioritize shaking costs, while compromising on spinning costs.
}
Both ABC-RL and AB-CPO perform nearly equally, and near the level of the baseline, though slightly worse. 
While its performance is retained, both shaking and spinning costs are greatly reduced with respect to the baseline. 
The spinning costs for both ABC-RL and AB-CPO are very close to 0, much lower than Const.
For shaking costs, AB-CPO has a lower curve than ABC-RL.

For comparison to human players, we also record 80 games in total played by 8 human players. The experimental results are shown as horizontal dashed lines in Figure \ref{fig: ml-agnet result-cpo}. 
The result shows that all the four agents perform better than human players and that the shaking costs of AB-CPO are comparable to human players, and the spinning costs of both ABC-RL and AB-CPO are close to zero, nearly the same as human players. This also shows that human players tend not to spin. 
Thus, the comparison shows that both ABC-RL and AB-CPO, particularly for AB-CPO, are able to retain a similar performance while behaving like humans in the aspects of shaking and spinning. 

\subsection{DMLab-30 Experiments}
\label{subsec:dmlab experiments}

\begin{figure}[htb!]
\centering
\medskip
{rooms\_keys\_doors\_puzzle}

\begin{minipage}{0.32\linewidth}
    \includegraphics[width=1\linewidth]{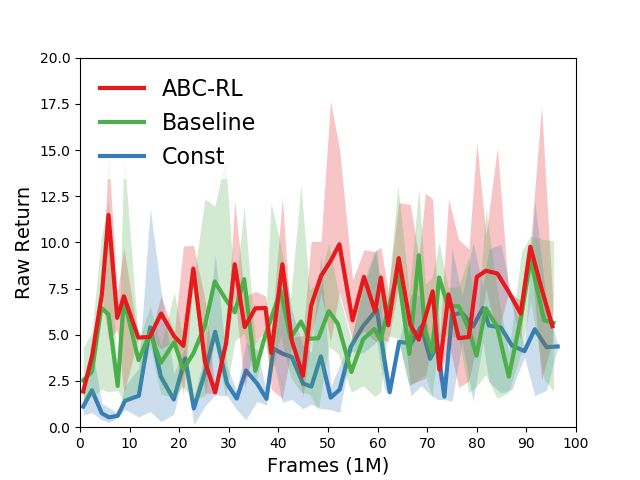}\\
    \centering
    return\label{fig: roomK_return}
\end{minipage}
\begin{minipage}{0.32\linewidth}
    \includegraphics[width=1\linewidth]{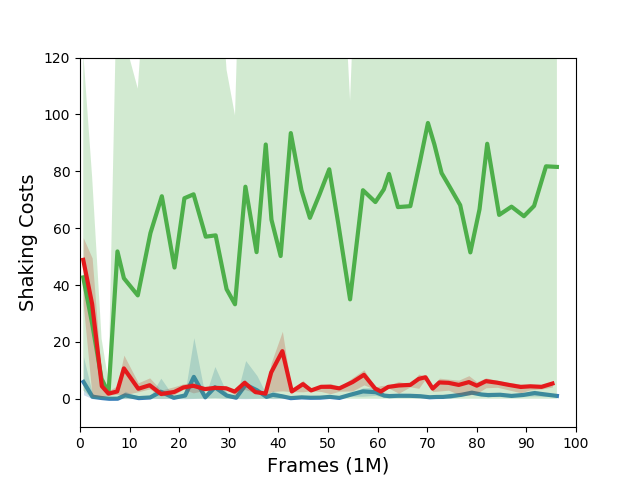}\\
    \centering
    shaking cost\label{fig: roomK_shake}
\end{minipage}
\begin{minipage}{0.32\linewidth}
    \includegraphics[width=1\linewidth]{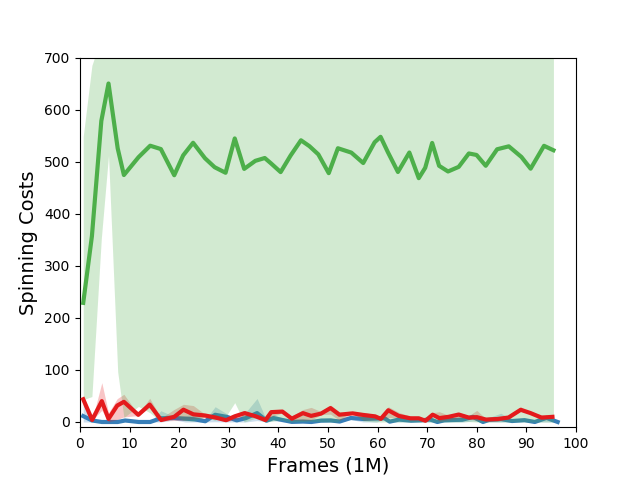}\\
    \centering
    spinning cost\label{fig: roomK_spin}
\end{minipage}

\medskip
{rooms\_collect\_good\_objects\_train}

\begin{minipage}{0.32\linewidth}
    \includegraphics[width=1\linewidth]{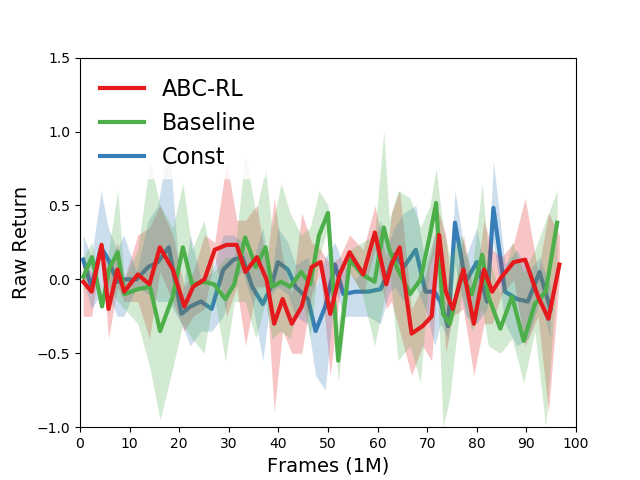}\\
    \centering
    return\label{fig: roomC_return}
\end{minipage}
\begin{minipage}{0.32\linewidth}
    \includegraphics[width=1\linewidth]{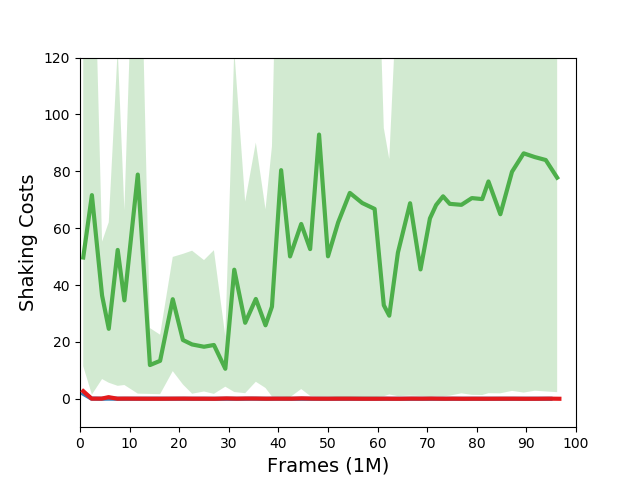}\\
    \centering
    shaking cost\label{fig: roomC_shake}
\end{minipage}
\begin{minipage}{0.32\linewidth}
    \includegraphics[width=1\linewidth]{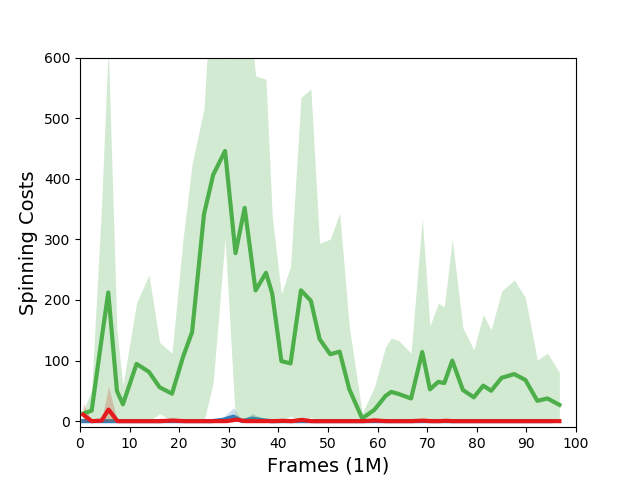}\\
    \centering
    spinning cost\label{fig: roomC_spin}
\end{minipage}


\medskip
{lasertag\_three\_opponents\_small}

\begin{minipage}{0.32\linewidth}
    \includegraphics[width=1\linewidth]{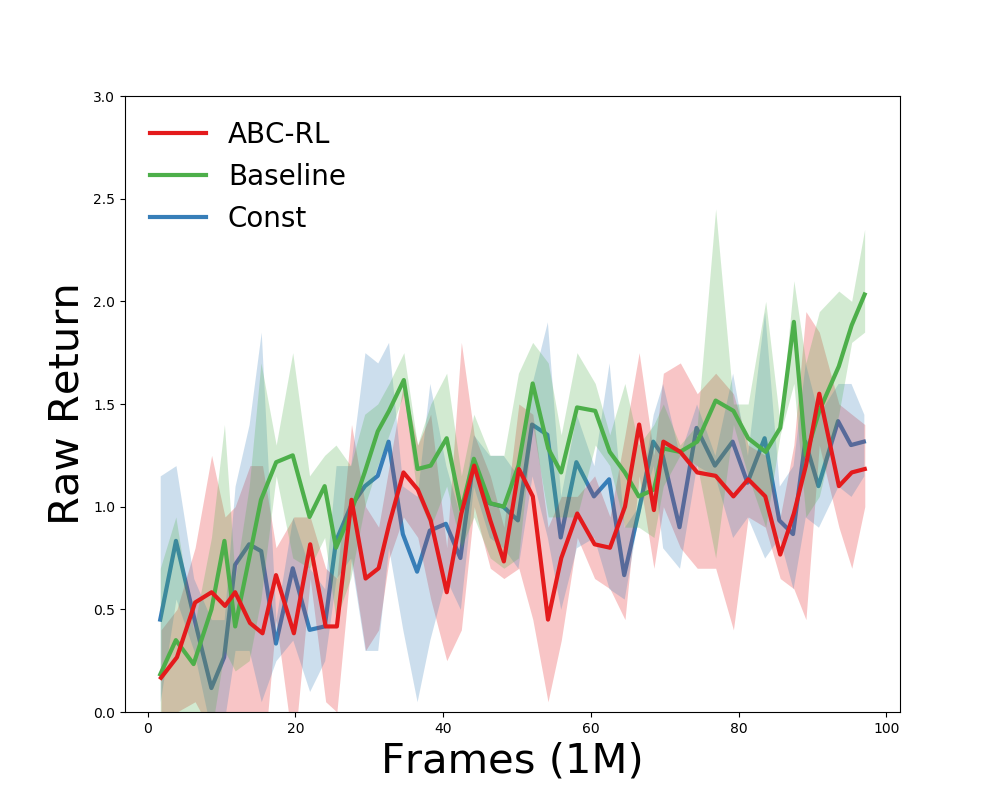}\\
    \centering
    return\label{fig: lasertag_return}
\end{minipage}
\begin{minipage}{0.32\linewidth}
    \includegraphics[width=1\linewidth]{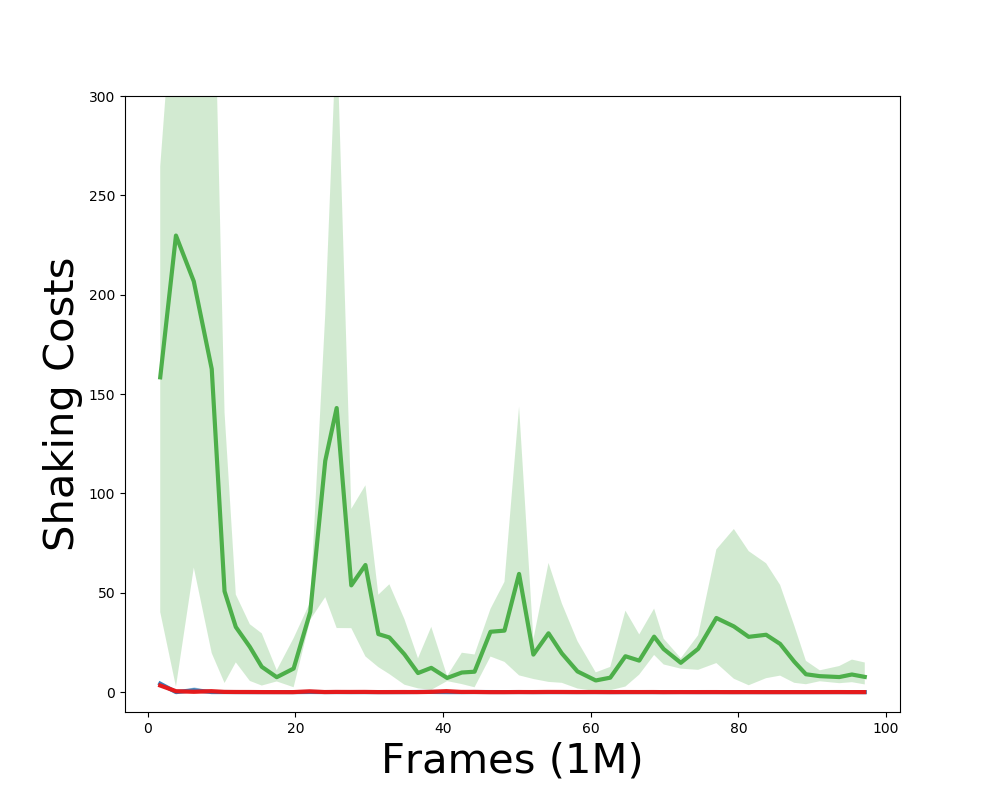}\\
    \centering
    shaking cost\label{fig: lasertag_shake}
\end{minipage}
\begin{minipage}{0.32\linewidth}
    \includegraphics[width=1\linewidth]{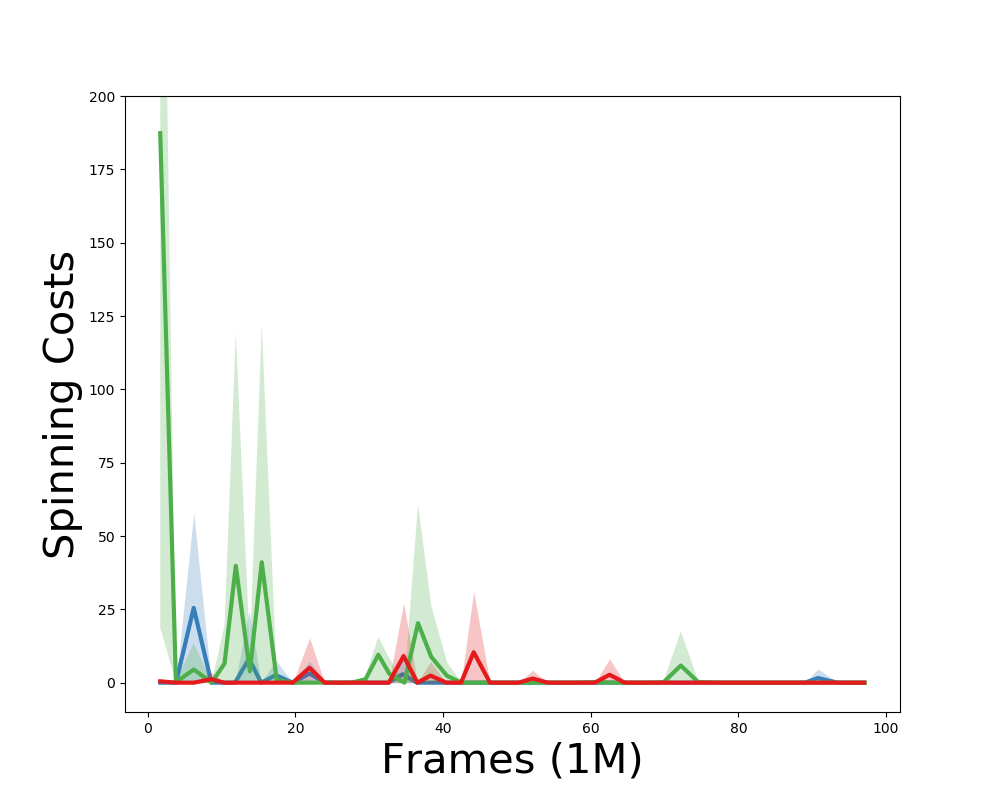}\\
    \centering
    spinning cost\label{fig: lasertag_spin}
\end{minipage}

\medskip
{rooms\_watermaze}

\begin{minipage}{0.32\linewidth}
    \includegraphics[width=1\linewidth]{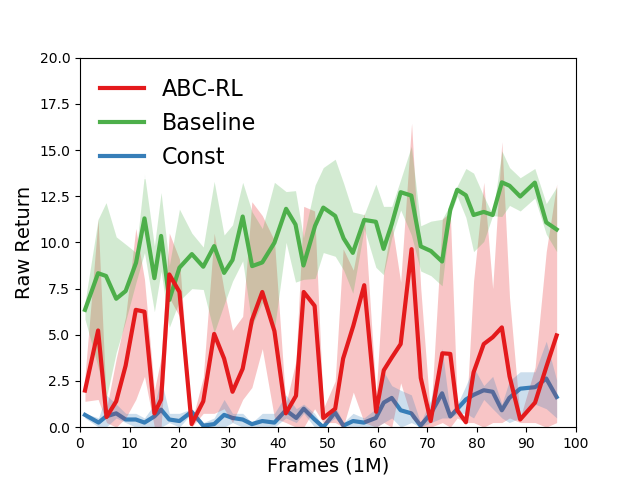}\\
    \centering
    return\label{fig: roomW_return}
\end{minipage}
\begin{minipage}{0.32\linewidth}
    \includegraphics[width=1\linewidth]{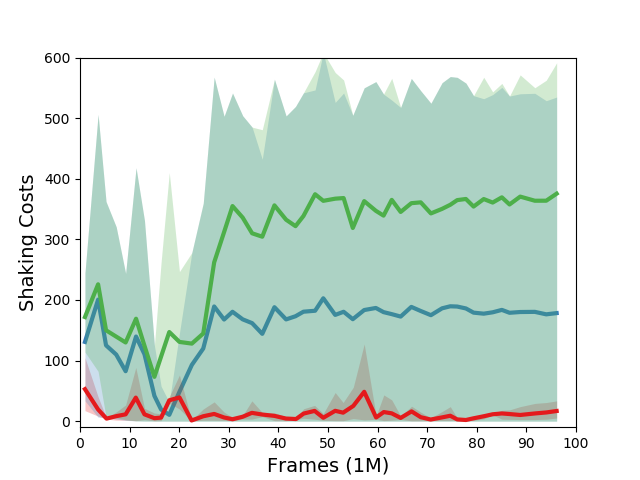}\\
    \centering
    shaking cost\label{fig: roomW_shake}
\end{minipage}
\begin{minipage}{0.32\linewidth}
    \includegraphics[width=1\linewidth]{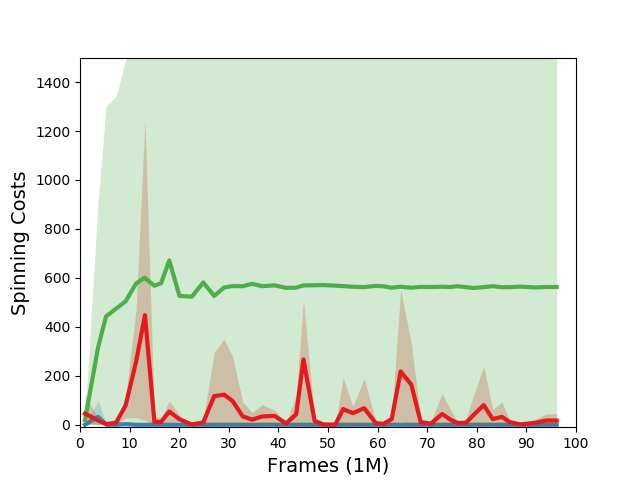}\\
    \centering
    spinning cost\label{fig: roomW_spin}
\end{minipage}


\caption{Results in DMLab-30 with training steps}
\label{fig: dmlab-30}
\end{figure}

\ho{
DMLab-30, provided by DeepMind Lab, is a collection of game environments for DRL research in first-person 3D world space.
In our study, we conducted experiments on four specific games from DMLab-30 as following:
\begin{enumerate}
    \item rooms\_collect\_good\_objects: In this game, the agent's objective is to collect good objects while avoiding bad objects. This game is similar to the Banana Collector game in the ML-Agents Toolkit.
    \item rooms\_keys\_doors\_puzzle: This game presents a procedural planning puzzle. The agent must navigate through a series of colored doors that block access to the goal object. The agent is able to keep only one single-use colored key to open the corresponding colored door. The challenge lies in determining the correct sequence of collecting keys and traversing rooms to reach the goal. 
    \item rooms\_watermaze: In this game, the agent's objective is to locate a hidden platform. Whenever the platform is discovered, the environment generates a reward for the agent and reset the agent's position. Finding the platform is challenging in initial trails, but the agent should remember its location in subsequent trials and navigate directly to it.
    \item lasertag\_three\_opponents\_small: In this game, the agent plays laser tag with other three opponent bots in a small map. The environment contains gadgets and power-ups, which are randomly generated. 
\end{enumerate}
These games provide diverse challenges and serve as a testbed for evaluating whether our proposed metrics and the training approach could be a general solution to the problem of creating human-like agents.
}


For DMLab-30, our experiments are based on the framework SEED RL\footnote{https://github.com/google-research/seed\_rl}. As SEED RL supports the implementation of v-trace, but not PPO, the version of 
\ds{V-trace}
serves as the baseline (without behavioral cost). With behavioral costs, both agents, ABC-RL and Const, are trained in the same way as those in 
\ds{Section~}\ref{subsec:ml-agents experiment}. Note that the proximity constraint Equation~\ref{eq:BCPO-opt-3 constraint} is not supported in 
\ds{V-trace}. So, we do not implement the agent AB-CPO for fairness of comparison mainly to the baseline, because ABC-RL is also representative as shown in the previous subsection. 
We run each experiment three times, each with around 100 million frames.

Figure \ref{fig: dmlab-30} shows the performances of the three agents, ABC-RL, baseline and Const, for the four games respectively. 
For the game, rooms\_keys\_doors\_puzzle, the performance of ABC-RL is comparable to the baseline, while both shaking and spinning costs are nearly zero, much lower than the baseline. 
It is similar for the game, rooms\_collect\_good\_objects. 
\ho{
For lasertag\_three\_opponents\_small, all methods are steadily improving in raw returns, but different in shaking and spinning costs.
ABC-RL and Const obtain near zero about shaking and spinining costs because they consider behavior costs.
Baseline shakes and spins a lot in the beginning, but after training, there is a slight amount of shaking and spinning left.
}
For the game, rooms\_watermaze, ABC-RL performs better than Const and worse than the baseline, while both shaking and spinning costs are also nearly zero, much lower than the baseline. 
However, it is interesting to observe the fluctuation of the performance of ABC-RL. 
Since agents in this game need to a find hidden platform for rewards\footnote{https://github.com/deepmind/lab/tree/master/game\_scripts/levels/contributed/dmlab30\#watermaze}, it is critical to take shaking and spinning actions. 
Once the agent ABC-RL receives high rewards, the behavioral weight grows, and hence it discourages the actions of shaking and spinning.
In such a situation, it becomes hard to find platforms and therefore the subsequent performance is reduced. 
The phenomenon illustrates the case that the behavioral costs are highly correlated to the performance increase. Whether there is a way of defining the cost to prevent this case is not in the scope of this paper.

\section{Conclusion}
\label{sec:conclusion}

The contribution of this paper is summarized as follows. 
First, we propose a new approach called Adaptive Behavioral Costs in Reinforcement Learning (ABC-RL) for training a human-like agent with competitive strength. 
To behave like humans and retain similar performance, ABC-RL augments behavioral limitations as cost signals in reinforcement learning with dynamically adjusted weights. 
Second, for ABC-RL, we propose a novel formulation that minimizes the behavioral costs subject to a constraint of the value function. 
By leveraging the augmented Lagrangian, our approach is an approximation of the Lagrangian adjustment, which handles the trade-off between the performance and the human-like behavior. 
Although this paper presents behavioral costs based on shaking and spinning, we leave the definition of the costs open, e.g., different $\alpha$ or other non-human-like actions. 
Third, in the 3D games of DMLab-30 and Unity ML-Agents Toolkit, our experiments show that ABC-RL preserves nearly the same performance level with significantly less shaking and spinning, as shown in Figures \ref{fig: ml-agnet result-cpo} and \ref{fig: dmlab-30}.



\bibliography{main} 

\end{document}